\documentclass[runningheads]{llncs}
\usepackage[T1]{fontenc}
\usepackage{graphicx}
\usepackage{amsmath}
\usepackage{amsfonts}
\usepackage{enumerate}
\usepackage[table]{xcolor}
\usepackage{tikz,pgfplots}
\usetikzlibrary{spy,calc, datavisualization.formats.functions, patterns}
\usetikzlibrary{positioning}
\pgfplotsset{compat=1.9}

\usepackage[colorlinks]{hyperref}
\usepackage[nameinlink,capitalize]{cleveref}
\usepackage{booktabs}
\usepackage{tabularx}
\usepackage{subcaption}
\usepackage{multirow}
\renewcommand{\paragraph}[1]{\medskip\noindent\textbf{#1}~~}

\usepackage{xspace}
\newcommand{\eg}{\textit{e.g.}\xspace}
\newcommand{\ie}{\textit{i.e.}\xspace}

\newcommand{\modelname}{NSVQ-GS\xspace}
\usepackage{mathrsfs}
\usepackage{amsmath,amsfonts,bm}

\def\eqref#1{equation~\ref{#1}}

\def\1{\bm{1}}

\def\vc{{\bm{c}}}

\def\ve{{\bm{e}}}

\def\vr{{\bm{r}}}
\def\vs{{\bm{s}}}
\def\vt{{\bm{t}}}

\def\vx{{\bm{x}}}

\def\vz{{\bm{z}}}

\def\mI{{\bm{I}}}
\def\mJ{{\bm{J}}}

\def\mR{{\bm{R}}}
\def\mS{{\bm{S}}}

\def\mW{{\bm{W}}}

\DeclareMathAlphabet{\mathsfit}{\encodingdefault}{\sfdefault}{m}{sl}
\SetMathAlphabet{\mathsfit}{bold}{\encodingdefault}{\sfdefault}{bx}{n}

\DeclareMathOperator*{\argmin}{arg\,min}

\newlength{\figurewidth}
\newlength{\figureheight}

\usepackage{xcolor}

\renewcommand{\orcidID}[1]{}
\begin{document}
\title{Compressing 3D Gaussian Splatting by Noise-Substituted Vector Quantization}
\titlerunning{Compressing 3D Gaussian Splatting by NSVQ}
\author{Haishan Wang\inst{1}\orcidID{0009-0002-0110-0161} \and
Mohammad Hassan Vali\inst{1}\orcidID{0000-0001-8023-6352} \and
Arno Solin\inst{1}\orcidID{0000-0002-0958-7886}}
\authorrunning{H. Wang et al.}
\institute{Department of Computer Science, Aalto University, Espoo, Finland\\
\email{\{haishan.wang, mohammad.vali, arno.solin\}@aalto.fi}}
\maketitle              %
\begin{abstract}
3D Gaussian Splatting (3DGS) has demonstrated remarkable effectiveness in 3D reconstruction, achieving high-quality results with real-time radiance field rendering. However, a key challenge is the substantial storage cost: reconstructing a single scene typically requires millions of Gaussian splats, each represented by 59 floating-point parameters, resulting in approximately 1~GB of memory. To address this challenge, we propose a compression method by building separate attribute codebooks and storing only discrete code indices. Specifically, we employ noise-substituted vector quantization technique to jointly train the codebooks and model features, ensuring consistency between gradient descent optimization and parameter discretization. Our method reduces the memory consumption efficiently (around $45\times$) while maintaining competitive reconstruction quality on standard 3D benchmark scenes. Experiments on different codebook sizes show the trade-off between compression ratio and image quality. Furthermore, the trained compressed model remains fully compatible with popular 3DGS viewers and enables faster rendering speed, making it well-suited for practical applications.

\end{abstract}

\section{Introduction}
\label{sec:intro}
\begin{figure*}[t!]
  \centering
  \setlength{\figurewidth}{0.24\textwidth}
  \foreach \method/\methodname/\size/\psnr/\fps [count=\j] in {
    {qualitative_half_thousand/bicycle_gt}/{Ground truth}/{}/{}/{},
    {qualitative_all_scenes/bicycle_3dgs}/{3DGS}/{1106}/{21.76}/{43},
    {qualitative_all_scenes/bicycle_newvq16k}/{ours (16k)}/{24.3}/{21.75}/{103},
    {qualitative_half_thousand/bicycle_newvq0.5k}/{ours (0.5k)}/{21.0}/{21.0}/{114}} {
    \begin{subfigure}[t]{\figurewidth}
    \begin{tikzpicture}[
      image/.style = {inner sep=0pt, outer sep=0pt, 
        minimum width=.95\figurewidth, anchor=north west, text width=.95\figurewidth}, 
      node distance = 1pt and 1pt, every node/.style={font= {\tiny}}, 
      label/.style = {font={\strut\footnotesize\bf},anchor=south,inner sep=0pt}, 
      spy using outlines={rectangle, magnification=1, size=0.45\figurewidth}
    ] 

      \node[image] (img-1) at (1*\figurewidth,0) 
        {\includegraphics[width=.95\figurewidth]{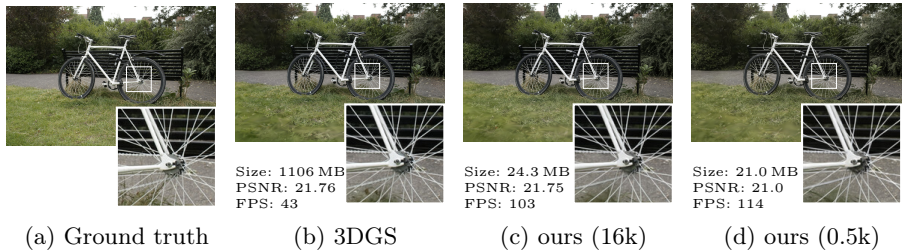}};

      \coordinate (spypoint-1-b) at ($(img-1.south east)+(0,.5cm)$);   

      \ifnum\j>1
        \node[inner sep=0,anchor=north west,align=left,font=\tiny] 
          at ($(img-1.south west) + (0,-1em)$)
          {Size: \size$\,$MB \\ PSNR: \psnr \\ FPS: \fps};
      \fi

      \coordinate (spot-1-b) 
        at ($(img-1.south west) + (0.6\figurewidth,.3\figurewidth)$); %
      \spy[white,magnification=4] on 
        (spot-1-b) in node[anchor=north east] at (spypoint-1-b);
    
    \end{tikzpicture}
    \caption{\methodname}
    \end{subfigure}%
  }
  \caption{We reduce the storage requirements by implementing an advanced VQ for 3DGS. It reduces file sizes and accelerates rendering speed, while maintaining high reconstruction quality. The reported frames per second (FPS) metrics were measured using an Nvidia RTX 4070 GPU.}
  \label{fig:teaser}
\end{figure*}

In computer graphics, 3D scene reconstruction has captured great attention from both academia and industry due to its wide range of applications. A key objective in this domain is novel view synthesis (NVS), which aims to generate novel images from new viewpoints based on a set of input images. Early approaches based on the multi-view stereo, such as structure-from-motion \cite{Schonberger_2016_CVPR_sfm}, provided robust and fundamental solutions to this task before the advent of deep learning.  Neural radiance field (NeRF, \cite{mildenhall2020nerf}) introduced neural networks to map spatial features to optical information. The latest advancement in this field is 3D Gaussian splatting (3DGS, \cite{kerbl20233d_3dgs}), which represents 3D scenes using a set of differentiable Gaussian primitives, often called splats. This technique significantly expands the boundaries of the domain by enabling high-fidelity reconstruction alongside real-time rendering, even for complex scenes. Consequently, 3DGS has been applied to various fields, including autonomous driving \cite{Zhou_2024_CVPR_DrivingGaussian}, AI-generated content \cite{Yi_2024_CVPR_GaussianDreamer}, Simultaneous Localization and Mapping (SLAM) \cite{Yan_2024_CVPR_gs_slam,Matsuki_2024_CVPR_gsslam}, and so on.
While 3DGS offers structure simplicity, computational efficiency and continuous rendering through explicit scene modelling, its high memory consumption remains a main limitation to further applications. Typically, representing a single scene requires millions of Gaussian splats, each characterized by 59 floating-point attributes, leading to substantial memory usage (\eg, approximately 1.1~GB for the {\sc bicycle} scene in the Mip-NeRF360 dataset; see \cref{fig:teaser}). Recent studies have revealed strong correlations between Gaussian attributes and a high dependency among Gaussian splats \cite{bagdasarian20243dgs_3dgszip}, indicating substantial information redundancy. These findings suggest the feasibility of employing compression techniques to reduce memory consumption with minimal impact on rendering performance.

Various 3DGS compression techniques have been proposed to reduce memory consumption while maintaining rendering quality. These approaches generally fall into two categories:
{\em (i)}~Machine learning (ML) methods, introduce hierarchical structures, predictive models, or neural networks to reduce redundancy. 
{\em (ii)}~Signal processing (SP) methods, which apply vector quantization, pruning, and entropy coding to optimize memory usage.
While ML-based methods offer strong compression, they introduce computational overhead due to their reliance on view-dependent neural networks and implicit representations, limiting applications requiring real-time rendering or explicit modelling. On the other hand, SP-based methods often struggle with optimization inconsistencies caused by the incompatibility between discretization and gradient descent. A key challenge remains: how to efficiently compress 3DGS while maintaining reconstruction quality, preserving GS advantages (real-time rendering speed and explicit scene modelling).

In this paper, we address this challenge by introducing {\bf \modelname}, a Noise-Substituted Vector Quantization (NSVQ, \cite{vali2022nsvq}) method that ensures optimization consistency while achieving an optimal balance between high compression ratios and high-fidelity reconstruction. Instead of treating quantization as a hard selection process, NSVQ models the quantization error by adding a noise term to the input vector such that it retains the statistical properties of the original error and thus enables direct optimization of the codebooks with gradient-based optimization.

Our main \textbf{contributions} are summarized as follows.
\begin{itemize}
  \item {\bf Compact 3DGS representation via NSVQ.} We introduce a discrete feature encoding method that maintains optimization consistency, avoid the clustering algorithms for code assignment, and achieve high compression ratios while preserving reconstruction quality across various bitrates.
  \item {\bf Efficient compression with real-time rendering and compatibility.} Our approach reduces memory usage and enables faster rendering, while keeping full compatibility with all existing 3DGS applications, such as web-based 3D visualization, 3D editing, and robotic vision.
  \item {\bf State-of-the-art performance.} We demonstrate state-of-the-art results on standard benchmarks in the category of signal processing (SP)-based GS compression, without reliance on any neural networks.
\end{itemize}

\section{Background and Related Work}
\label{sec:background}
We provide the necessary background on 3DGS, focusing on its parameter structure and rendering process, then review the existing GS compression approaches.

\subsection{3D Gaussian Splatting}
\label{subsec:bg_gs}
The 3D scene is modelled by 3DGS as a set of Gaussian splats. Each Gaussian splat consists of 6 attributes: 3D spatial coordinates $\vx \in \mathbb{R}^3$, opacity $o \in \mathbb{R}$, scaling and rotation parameters $\vs \in \mathbb{R}^3, \vr \in \mathbb{R}^4$ which jointly represent the covariance matrix $\Sigma = \mR\mS\mS^\top \mR^\top \in \mathbb{R}^{3\times 3}$, colours $\vc \in \mathbb{R}^{3}$ and spherical harmonics (SH) coefficients $\vc^{sh} \in \mathbb{R}^{45}$ of order 3 (the dimensions of SH depend on the order, 3\textsuperscript{rd} order is the convention trade-off between performance and cost). The scaling and rotation matrices $\mS, \mR$ are recovered by corresponding parameters $\vs, \vr$. The pixel-wise colour rendering of 3DGS keeps the image formation of pixel-based $\alpha$-blending and volumetric rendering in NeRF \cite{kerbl20233d_3dgs}. The pixel colour $C$ is calculated by $\alpha$-blending: 
\begin{equation}
  \label{eq:alpha-blending}
  C = \sum_{i=1}^{|\mathcal{N}|}\tilde{\vc}_i \alpha_i \prod_{j=1}^{n-1} (1-\alpha_j),
\end{equation}
where $\mathcal{N}$ refers to all  Gaussians splats visible from the viewpoint of the current pixel, which are sorted by depth, $\tilde{\vc}_i$ denotes the colour recovered from colours $\vc_i$ and spherical harmonics $\vc_i^{sh}$, and $\alpha_i$ is the alpha blending term obtained by scaling the opacity by the Gaussian distribution
$$
\alpha_i = o_i \exp\left({-\frac{1}{2}(\vx'-\mu_i')\Sigma_i'^{-1} (\vx'-\mu_i')^\top}\right),
$$
where $\vx', \mu_i'$ denote the projected coordinates of the pixel and the Gaussian splat. The covariance matrix after 2D-projection is $\Sigma' = \mJ \mW \Sigma \mW^\top \mJ^\top$, where $\mJ, \mW$ denote the Jacobian of the affine approximation of projection and viewing transformation.

\subsection{Compression Approaches for 3DGS}
\label{subsec:bg_compress_gs}
3DGS has achieved remarkable success in the 3D reconstruction domain with a wide range of applications. However, the high storage costs limit its widespread adoption. Over the past few years, researchers have developed many methods to address this limitation, which can be classified into two fundamental strategies: \textbf{compaction}, which reduces the number of splats through adaptive density control (ADC) and improved heuristic, and \textbf{compression}, which optimizes the organization of attributes to minimize redundancy. This paper focuses on developing a GS compression method that is compatible with most existing compaction techniques.

Existing compression approaches can be categorized into two types: Signal processing-based (\textbf{SP-based}) and machine learning-based (\textbf{ML-based}). 

\paragraph{SP-based methods} often employ techniques such as vector quantization (VQ), which discretize high-dimensional continuous feature spaces into a set of representative codewords \cite{fan2023lightgaussian,Niedermayr_2024_CVPR_compressed3dgs,10.1145/3651282_reduced3dgs,navaneet2024compgs}. Besides VQ techniques, LightGaussians \cite{fan2023lightgaussian} adaptively distils SH parameters and improves ADC by removing Gaussian splats with minimal global significance of reconstruction. Compressed3D \cite{Niedermayr_2024_CVPR_compressed3dgs} introduces space-filling curves for efficient coordinates information storage. Reduced3DGS \cite{10.1145/3651282_reduced3dgs} estimates splat redundancy in a scale- and resolution-aware manner for ADC, selects SH bands adaptively, and suggests half-floating data representation. CompGS~\cite{navaneet2024compgs} utilizes periodic K-means clustering for codebook assignment and incorporates an opacity regularization to control splats amount.

Despite these advancements, the issue of {\em gradient collapse} \cite{vali2025vector} in VQ-based methods has not received sufficient attention. For example, Reduced3DGS avoids training on VQ, while LightGaussians and Compressed3D fix the codebook assignment during training. CompGS addresses this issue using a straight-through estimator (STE) which copies the gradients through VQ function. These limitations motivate us to propose \modelname.

\paragraph{ML-based methods} leverage techniques from the machine learning domain, such as Self-organizing Maps \cite{morgenstern2024compact_SOG,ye2024gsplat} or hash grids \cite{chen2025hac}. Some methods utilize simple multilayer perceptrons (MLPs) as decoders for Gaussian attributes, particularly for colour-related features \cite{girish2024eagles,Lee_2024_CVPR_compact3dgs}. A prominent example of ML-based methods is Scaffold-GS \cite{lu2024scaffold}, which introduces anchor points for the hierarchical structure of GS attributes. For each anchor, the model reconstructs a group of neighbouring splats by low-dimensional embeddings via several shared view-dependent decoders. This idea has inspired several follow-up works. For instance, HAC \cite{chen2025hac} designs an adaptive quantization module on anchor attribute values, and predicts anchor attributes by querying anchor coordinates in the hash grid. ContextGS \cite{wang2024contextgs} refines anchors reconstruction from coarse to fine granularity using autoregressive models with quantized hyperpriors.

\begin{figure}[t!]
  \resizebox{\textwidth}{!}{%
  \begin{tikzpicture}[remember picture]

    \pgfdeclarelayer{background}
    \pgfsetlayers{background,main} %

    \newlength{\boxsize}
    \setlength{\boxsize}{1em}

    \tikzset{mybox/.style={fill=white, draw=black!50!blue, inner sep=0, outer sep=0, minimum width=\boxsize, minimum height=\boxsize,rounded corners=1pt}}
    \tikzset{
      star hatch/.style={pattern=north west lines, pattern color=black},
      diagonal hatch/.style={pattern=north east lines, pattern color=black},
      grid hatch/.style={pattern=grid, pattern color=black},
      dotted hatch/.style={pattern=dots, pattern color=black}
    }

    \tikzset{dbox/.style={fill=white, draw=black!50!blue, inner sep=0, outer sep=0,rounded corners=1pt}
    }

    \newcommand{\dynamicbox}[4]{%
      \node[dbox, minimum width={#1*\boxsize}, minimum height={#2*\boxsize}] (#3) at (#4) {};%
    }
    
    \newcommand{\boxes}[3]{%
      \foreach \i in {1,...,#1} { 
        \node[mybox] (#2-\i) at ($#3 + (\boxsize*\i,0)$) {};
      }
    }
    \newcommand{\moreboxes}[3]{%
      \foreach \j in {1,...,5}
        \foreach \i in {1,...,#1} { 
            \node[mybox] (#2-\j-\i) at ($#3 + (\boxsize*\i,\boxsize*\j)$) {};
        } 
    }    

    \node[anchor=north west,align=center] at (0,0) {
    \strut \textbf{Gaussians}\\[1.2em]
    \setlength{\tabcolsep}{3pt}
    \begin{tabular}{lll}
      $\vx$ & \tikz[remember picture,baseline=-2ex]{\boxes{3}{x}{(0,0)}} & $\mathbb{R}^3$ \\
      $o$   & \tikz[remember picture,baseline=-2ex]{\boxes{1}{o}{(0,0)}} & $\mathbb{R}$ \\
      $\vs$ & \tikz[remember picture,baseline=-2ex]{\boxes{3}{s}{(0,0)}} & $\mathbb{R}^3$ \\
      $\vr$ & \tikz[remember picture,baseline=-2ex]{\boxes{4}{r}{(0,0)}} & $\mathbb{R}^4$ \\
      $\vc$ & \tikz[remember picture,baseline=-2ex]{\boxes{3}{c}{(0,0)}} & $\mathbb{R}^3$ \\
      $\vc^{sh}$ & \tikz[remember picture,baseline=10ex]{\moreboxes{9}{sh}{(0,0)}} & $\mathbb{R}^{45}$ \\
      
    \end{tabular}
    };

    \node[anchor=north west,align=center] at (5,0) {
    \strut \textbf{Quantized Gaussians}\\[1.2em]
    \setlength{\tabcolsep}{3pt}
    \begin{tabular}{lll}
      $\vx$ & \tikz[remember picture,baseline=-2ex]{\boxes{3}{qx}{(0,0)}} & $\mathbb{R}^3$ \\
      $o$   & \tikz[remember picture,baseline=-2ex]{\boxes{1}{qo}{(0,0)}} & $\mathbb{R}$ \\
      $k_{\vs}^*$ & \tikz[remember picture,baseline=-2ex]{\dynamicbox{14/32}{1}{qs-1}{0,0}}~\textcolor{red}{\tiny 14 bits} & $\{1,\ldots,2^{K_s}\}$ \\
      $k_{\vr}^*$ & \tikz[remember picture,baseline=-2ex]{\dynamicbox{14/32}{1}{qr-1}{0,0}}~\textcolor{red}{\tiny 14 bits} & $\{1,\ldots,2^{K_r}\}$ \\
      $k_{\vc}^*$ & \tikz[remember picture,baseline=-2ex]{\dynamicbox{12/32}{1}{qc-1}{0,0}}~\textcolor{red}{\tiny 12 bits} & $\{1,\ldots,2^{K_c}\}$ \\
      $k_{\vc^{sh}}^*$ & \tikz[remember picture,baseline=-2ex]{\dynamicbox{12/32}{1}{qsh-1}{0,0}}~\textcolor{red}{\tiny 12 bits} & $\{1,\ldots,2^{K_{sh}}\}$ \\
      
    \end{tabular}
    };    

    \node[anchor=north west,align=center, inner sep=3pt,rounded corners=1pt,draw=black!80] (VQ) at (7,-5) {\large VQ};

    \node[anchor=north west,align=center] at (11,0) {
      \strut \textbf{Codebooks}};

    \node (p) at (13,-1) {};
    \foreach \book/\bs [count=\b] in {{$\mathcal{C}_{sh}$}/7,{$\mathcal{C}_c$}/3,{$\mathcal{C}_r$}/4,{$\mathcal{C}_s$}/3} {
      \draw[draw=black,fill=black!05,rounded corners=1pt] ($(p) - (0.5cm*\b,0)$) -- ++(0,-1.5cm) -- ++(\bs*6pt,-\bs*3pt) -- ++(0,1.5cm) -- cycle;
      
      \foreach \i in {1,...,\bs}
        \draw[black,opacity=.3] ($(p) - (0.5cm*\b,0) + (\i*6pt,-\i*3pt)$) -- ++(0,-1.5cm);
       
      \node[align=center] at ($(p) - (0.5cm*\b,-.25cm)$) {\book};
    }

    \node[align=center,font=\tiny,anchor=east] at ($(p) - (0.5cm*4,.75cm)$) {1 \\[6pt] $\vdots$ \\[6pt] $2^{K_s}$};

    \node[anchor=north west,align=left] (legend) at (10,-4) {
    \strut \textbf{Legend}\\[.5em]
    \setlength{\tabcolsep}{3pt}
    \begin{tabular}{ll}
      \tikz[remember picture,baseline=-2ex]{\boxes{1}{ls}{(0,0)}} & Scaling \\
      \tikz[remember picture,baseline=-2ex]{\boxes{1}{lr}{(0,0)}} & Rotation \\
      \tikz[remember picture,baseline=-2ex]{\boxes{1}{lc}{(0,0)}} & Colour \\
      \tikz[remember picture,baseline=-2ex]{\boxes{1}{lsh}{(0,0)}} & Spherical harmonics \\      
    \end{tabular}
    };     

    \node[mybox,fill=red] at (x-1) {};
    \node[font=\tiny,red,anchor=south] at (x-1.north) {32 bits};

    \foreach \i in {1,...,3} {
      \node[mybox,fill=red,star hatch] at (s-\i) {};
    }
    \foreach \i in {1,...,4} {
      \node[mybox,fill=red,grid hatch] at (r-\i) {};
    }
    \foreach \i in {1,...,3} {
      \node[mybox,fill=red,diagonal hatch] at (c-\i) {};
    }
    \foreach \i in {1,...,5} 
      \foreach \j in {1,...,9} {
        \node[mybox,fill=red,dotted hatch] at (sh-\i-\j) {};
    }
    \node[dbox,fill=red,star hatch, minimum width=14/32*\boxsize, minimum height=1*\boxsize] at (qs-1) {};
    \node[dbox,fill=red,grid hatch, minimum width=14/32*\boxsize, minimum height=\boxsize] at (qr-1) {};
    \node[dbox,fill=red,diagonal hatch, minimum width=12/32*\boxsize, minimum height=\boxsize] at (qc-1) {};
    \node[dbox,fill=red,dotted hatch, minimum width=12/32*\boxsize, minimum height=\boxsize] at (qsh-1) {};

    \node[mybox,fill=red,star hatch] at (ls-1) {};
    \node[mybox,fill=red,grid hatch] at (lr-1) {};
    \node[mybox,fill=red,diagonal hatch] at (lc-1) {};
    \node[mybox,fill=red,dotted hatch] at (lsh-1) {};

    \begin{pgfonlayer}{background}
      \node[fill=blue!05,anchor=north west,rounded corners, dashed, draw=blue!20,
          minimum width=4.6cm,minimum height=3cm] (fill-1)
          at ($(s-1) + (-1cm,6pt)$) {};
      \node[fill=blue!05,anchor=north west,rounded corners, dashed, draw=blue!20,
          minimum width=4.3cm,minimum height=1.8cm] (fill-2)
          at ($(qs-1) + (-1cm,6pt)$) {};
      \node[fill=blue!05,anchor=north west,rounded corners, dashed, draw=blue!20,
          minimum width=4.3cm,minimum height=3cm+2pt] (fill-3)
          at ($(p) + (-2.8cm,16pt)$) {};
                    
      \draw[line width=2pt,<-,draw=blue!20] (VQ) -| (fill-1); 
      \draw[line width=2pt,<-,draw=blue!20] (VQ) -| (fill-3);    
      \draw[line width=2pt,->,draw=blue!20] (VQ) -- (fill-2); 

      \fill[fill=white,opacity=.75,draw=black!10,rounded corners] (legend.north west) rectangle (legend.south east);    
      
    \end{pgfonlayer}
  
  \end{tikzpicture}}
  \caption{Overview of the efficient reduction on storage requirement by our \modelname~(16k). A single unit box represents 32 bits. Substituting Gaussian splats with their quantized counterparts and codebooks saves substantial memory consumption.} 
  \label{fig:quantization}
\end{figure}
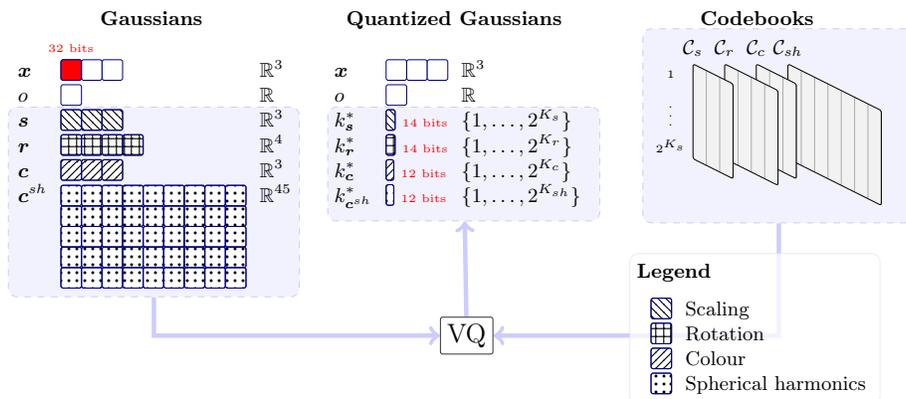

\section{Methods}
\label{sec:methods}
In this section, we explain the NSVQ technique \cite{vali2022nsvq} that optimizes VQ codebook by gradient-based optimizers and then, we present the training process of our proposed method \modelname.
\subsection{Noise Substitution in Vector Quantization}
\label{subsec:method_nsvq}
Vector quantization (VQ, \cite{gersho1992vector}) is a classical signal processing technique that is used to compress a continuous data distribution with a limited discrete set of representative vectors called a codebook. Each codebook vector represents a subset of the data distribution, such that it is the closest codebook vector to all data samples in the subset. Given an input $\vt \in \mathbb{R}^{1\times D}$ and a codebook $\mathcal{C} = \{\vz_k\in\mathbb{R}^{1\times D} \mid k\in{1,\dots,2^K}\} \in \mathbb{R}^{2^K\times D}$ of bitrates $K\in \mathbb{N}$, the hard quantized input $\vt_q$ is computed as
 \begin{equation}
  \label{eq:hard_quantization}
 \vt_q = \vz_{k^*}, \quad k^* = \argmin_{k \in \{1,\dots,2^K\} } \|\vt - \vz_k\|_2, 
\end{equation}
where $k^*$ is the index of the closest code from $\mathcal{C}$ to the input $\vt$ and ${\|\cdot\|}_2$ refers to the Euclidean distance.

According to \cref{eq:hard_quantization}, the VQ is nondifferentiable and therefore cannot propagate gradients during the backward pass in neural network training. This issue, known as the \textit{gradient collapse} problem \cite{vali2025vector}, prevents effective learning. A common approach used to address this problem is the straight-through estimator (STE, \cite{bengioSTE}), which copies the gradients unchanged over the VQ function during backpropagation. Despite its simplicity, STE introduces several limitations: it incurs additional hyper-parameter tuning, modifies the optimization hyperplane due to the inclusion of supplementary loss terms in the training objective, and fails to account for quantization effects during training.

Noise substitution in vector quantization (NSVQ, \cite{vali2022nsvq}) is another solution to \textit{gradient collapse} that leads to faster convergence, more accurate gradients, and less hyper-parameter tuning than STE. NSVQ simulates the quantization by adding noise to the input vector such that the noise has the original quantization magnitude but in a random direction. NSVQ quantizes a given input vector $\vt$ as
\begin{equation}
  \label{eq:nsvq}
\tilde{\vt}_q = \vt + {\|\vt-  \vt_q\|}_2\cdot \frac{\ve}{\|\ve\|_2}, \quad \ve \sim \mathrm{N}(\bm{0},\mI),
\end{equation}
where $\vt_q=\vz_{k^*}$ is the hard quantized version of input (see \cref{eq:hard_quantization}). Since $\tilde{\vt}_q$ is a differentiable function of input $\vt$ and selected codebook vector of $\vz_{k^*}$, it can be used directly in end-to-end training of neural networks to backpropagate gradients through non-differentiable VQ function.

{\em Codebook collapse} \cite{vali2025vector,dieleman2018challenge} is a common challenge in the training of VQ codebooks, where a subset of codebook vectors remain unused for quantization. As a result, these codebook vectors are not updated and remain inactive throughout training. To address this challenge, we adopt the {\em codebook replacement} procedure proposed in \cite{vali2022nsvq}, \ie, after a predefined number of training batches, inactive codebook vectors, those used less than a threshold are replaced with a permutation of a randomly selected set of actively used ones.

\subsection{Proposed Method}
\label{sec:proposed}
The memory consumption of 3DGS arises mainly from the substantial amount of splats and associated attributes. For an efficient representation of them, we employ NSVQ to quantize four Gaussian attributes: colours, SH, scaling and rotation parameters. This approach strikes an optimal balance between compression efficiency (memory reduction) and minimal additional degradation to reconstruction quality. During model storage and rendering, the quantized features are modeled by the codebooks and the corresponding indices. For instance, with setting codebook bitrates as $10$, the $45$-dimensional SH features, originally stored as 32-bit floating-point values requiring 1,440 bytes, can be replaced by a single index requiring only 1.25 bytes. The detailed training process of proposed \modelname is described below.

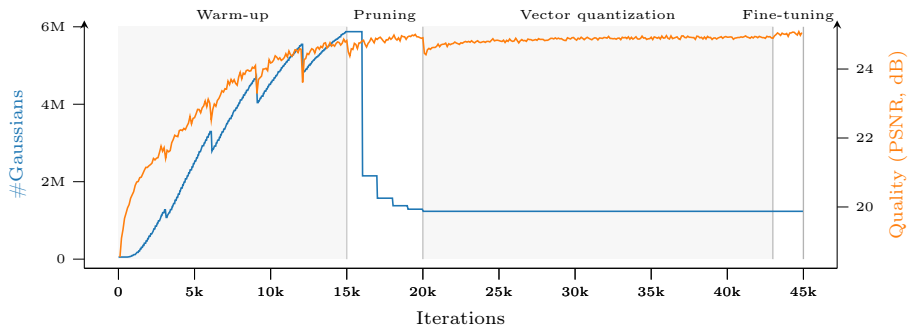
\begin{figure}[t!]
  \centering\scriptsize
  \setlength{\figurewidth}{.95\textwidth}
  \setlength{\figureheight}{.4\textwidth}
  \pgfplotsset{
    y tick label style={font=\tiny},
    x tick label style={font=\tiny},
    xtick={0,5000,10000,15000,20000,25000,30000,35000,40000,45000},
    xticklabels={0,5k,10k,15k,20k,25k,30k,35k,40k,45k},
    scaled ticks=false
  }
\begin{tikzpicture}

\definecolor{darkgrey176}{RGB}{176,176,176}
\definecolor{darkorange25512714}{RGB}{255,127,14}
\definecolor{lightgrey204}{RGB}{204,204,204}
\definecolor{steelblue31119180}{RGB}{31,119,180}

\colorlet{shadecolor}{black!10}

\newcommand{\shading}[4]{%
  \addplot [fill=#1, opacity=0.3, draw=none] coordinates {
    (#2, 0) (#2, 6000000) (#3, 6000000) (#3, 0)
  } --cycle;

  \draw[black!30,line width=.5pt] (axis cs: #3, 0) -- (axis cs: #3,6000000);

  \node[rotate=0, anchor=south, font=\tiny\strut, inner sep=1pt] 
    at ($(axis cs: #2, 6000000)!.5!(axis cs: #3, 6000000)$) {#4};
    
}

\begin{axis}[
clip=false,
axis y line=left,
axis x line*=bottom,
height=\figureheight,
legend cell align={left},
legend style={
  fill opacity=0.8,
  draw opacity=1,
  text opacity=1,
  at={(0.91,0.5)},
  anchor=east,
  draw=lightgrey204
},
tick align=outside,
tick pos=left,
width=\figurewidth,
x grid style={darkgrey176},
xlabel={Iterations},
xmin=-2242.15, xmax=47217.15,
xtick style={color=black},
y grid style={darkgrey176},
ylabel={\#Gaussians},
ymin=-236719.45, ymax=6165158.45,
ytick style={color=black},
ytick={0,2000000,4000000,6000000},
yticklabels={0,2M,4M,6M},
every axis y label/.append style ={steelblue31119180},
]

\shading{shadecolor}{0}{15000}{Warm-up}
\shading{white}{15000}{20000}{Pruning}
\shading{shadecolor}{20000}{43000}{Vector quantization}
\shading{white}{43000}{45000}{Fine-tuning}

\addplot [semithick, steelblue31119180,forget plot]
table {%
6 54275
40 54275
82 54275
122 54275
161 54275
204 54275
233 54275
306 54275
342 54275
396 54275
437 54275
497 54275
552 54275
598 54275
644 58437
692 58437
753 67702
794 67702
848 80166
876 80166
906 96231
958 96231
1014 117281
1070 117281
1097 117281
1155 143351
1223 175529
1300 175529
1336 213503
1392 213503
1422 256470
1468 256470
1504 303669
1574 303669
1613 354792
1636 354792
1674 354792
1702 410889
1753 410889
1785 410889
1818 468270
1881 468270
1924 528563
1964 528563
2006 590645
2039 590645
2083 590645
2124 654587
2190 654587
2276 717803
2334 783831
2392 783831
2426 852253
2468 852253
2497 852253
2548 923269
2586 923269
2631 990398
2698 990398
2766 1059462
2819 1128888
2855 1128888
2904 1200091
2938 1200091
2992 1200091
3037 1274469
3083 1274469
3136 1077474
3188 1077474
3230 1134147
3334 1204272
3370 1204272
3417 1280110
3457 1280110
3490 1280110
3532 1357675
3563 1357675
3627 1437550
3661 1437550
3695 1437550
3717 1518614
3743 1518614
3791 1518614
3833 1595802
3891 1595802
3931 1677169
3961 1677169
3998 1677169
4038 1757455
4064 1757455
4115 1836026
4163 1836026
4188 1836026
4225 1914038
4258 1914038
4315 1990109
4367 1990109
4405 2074757
4477 2074757
4528 2151549
4575 2151549
4620 2228187
4669 2228187
4712 2305591
4763 2305591
4796 2305591
4835 2384097
4889 2384097
4927 2464086
4971 2464086
5024 2536463
5083 2536463
5136 2617264
5186 2617264
5229 2689672
5261 2689672
5307 2774464
5349 2774464
5384 2774464
5444 2846716
5509 2928987
5550 2928987
5595 2928987
5638 2999026
5676 2999026
5733 3072937
5786 3072937
5818 3148819
5850 3148819
5893 3148819
5934 3223895
5975 3223895
6019 3294261
6061 3294261
6099 3294261
6144 2790222
6210 2846983
6242 2846983
6306 2919467
6349 2919467
6388 2919467
6424 2995225
6461 2995225
6519 3071145
6558 3071145
6596 3071145
6634 3142158
6686 3142158
6718 3212759
6752 3212759
6816 3288547
6862 3288547
6883 3288547
6922 3358289
6984 3358289
7057 3432218
7118 3500845
7161 3500845
7210 3566529
7253 3566529
7307 3630696
7350 3630696
7396 3630696
7439 3702820
7478 3702820
7529 3775354
7607 3840456
7644 3840456
7682 3840456
7723 3900957
7750 3900957
7785 3900957
7834 3967016
7876 3967016
7926 4034277
7965 4034277
8009 4110568
8067 4110568
8090 4110568
8149 4172622
8197 4172622
8223 4238973
8267 4238973
8294 4238973
8341 4298901
8388 4298901
8418 4357977
8441 4357977
8496 4357977
8563 4418820
8622 4481427
8669 4481427
8738 4535620
8781 4535620
8829 4597857
8877 4597857
8919 4652327
8967 4652327
8999 4652327
9061 4714786
9122 4042828
9141 4042828
9177 4042828
9221 4088581
9252 4088581
9310 4152095
9353 4152095
9402 4208500
9445 4208500
9494 4208500
9522 4274093
9557 4274093
9603 4338481
9649 4338481
9683 4338481
9727 4396406
9773 4396406
9822 4456984
9856 4456984
9902 4513614
9945 4513614
9986 4513614
10022 4566484
10067 4566484
10117 4619418
10145 4619418
10189 4619418
10233 4681740
10260 4681740
10292 4681740
10345 4738955
10396 4738955
10461 4790372
10502 4840223
10549 4840223
10590 4840223
10641 4885356
10673 4885356
10709 4939648
10754 4939648
10811 4986562
10860 4986562
10916 5036346
10955 5036346
11030 5091872
11079 5091872
11113 5136759
11150 5136759
11193 5136759
11218 5197190
11266 5197190
11301 5233088
11371 5233088
11396 5233088
11446 5279960
11499 5279960
11536 5329828
11588 5329828
11658 5368215
11712 5418687
11748 5418687
11793 5418687
11825 5455506
11875 5455506
11904 5507330
11938 5507330
11970 5507330
12020 5546953
12070 5546953
12097 5546953
12131 4807896
12217 4852332
12252 4852332
12306 4902387
12341 4902387
12378 4902387
12418 4953808
12458 4953808
12501 5004642
12541 5004642
12582 5004642
12608 5046142
12664 5046142
12713 5098686
12743 5098686
12812 5138261
12853 5138261
12911 5183555
12952 5183555
13011 5222872
13084 5222872
13125 5258163
13177 5258163
13234 5301867
13278 5301867
13319 5337095
13356 5337095
13408 5380107
13477 5380107
13506 5411814
13557 5411814
13602 5449900
13650 5449900
13701 5485806
13744 5485806
13781 5485806
13814 5517869
13854 5517869
13893 5517869
13938 5557151
13984 5557151
14029 5591586
14056 5591586
14104 5624215
14161 5624215
14221 5653668
14268 5653668
14314 5682646
14357 5682646
14391 5682646
14444 5716887
14477 5716887
14528 5741731
14556 5741731
14597 5741731
14631 5789570
14671 5789570
14724 5815843
14760 5815843
14792 5815843
14858 5843106
14891 5843106
14929 5874164
14971 5874164
15017 5874164
15093 5874164
15136 5874164
15180 5874164
15236 5874164
15267 5874164
15297 5874164
15338 5874164
15378 5874164
15419 5874164
15474 5874164
15522 5874164
15554 5874164
15584 5874164
15671 5874164
15724 5874164
15772 5874164
15819 5874164
15863 5874164
15908 5874164
15951 5874164
15995 5874164
16038 2150785
16108 2150785
16148 2150785
16186 2150785
16222 2150785
16257 2150785
16293 2150785
16349 2150785
16392 2150785
16426 2150785
16461 2150785
16499 2150785
16553 2150785
16588 2150785
16644 2150785
16679 2150785
16740 2150785
16772 2150785
16802 2150785
16844 2150785
16889 2150785
16934 2150785
16977 2150785
17013 1573977
17052 1573977
17094 1573977
17140 1573977
17190 1573977
17248 1573977
17302 1573977
17364 1573977
17410 1573977
17485 1573977
17536 1573977
17562 1573977
17617 1573977
17659 1573977
17689 1573977
17724 1573977
17778 1573977
17834 1573977
17871 1573977
17918 1573977
17953 1573977
18000 1573977
18041 1379412
18093 1379412
18129 1379412
18186 1379412
18222 1379412
18269 1379412
18314 1379412
18390 1379412
18433 1379412
18497 1379412
18533 1379412
18557 1379412
18591 1379412
18631 1379412
18671 1379412
18700 1379412
18740 1379412
18806 1379412
18859 1379412
18893 1379412
18928 1379412
18976 1379412
19024 1289611
19070 1289611
19103 1289611
19147 1289611
19209 1289611
19246 1289611
19285 1289611
19358 1289611
19398 1289611
19442 1289611
19486 1289611
19515 1289611
19547 1289611
19586 1289611
19626 1289611
19693 1289611
19726 1289611
19760 1289611
19789 1289611
19830 1289611
19889 1289611
19935 1289611
19960 1289611
20021 1234897
20064 1234897
20100 1234897
20146 1234897
20184 1234897
20248 1234897
20284 1234897
20325 1234897
20353 1234897
20405 1234897
20445 1234897
20491 1234897
20540 1234897
20575 1234897
20612 1234897
20662 1234897
20703 1234897
20739 1234897
20782 1234897
20851 1234897
20928 1234897
20980 1234897
21044 1234897
21078 1234897
21125 1234897
21157 1234897
21190 1234897
21228 1234897
21261 1234897
21298 1234897
21335 1234897
21362 1234897
21414 1234897
21468 1234897
21527 1234897
21555 1234897
21601 1234897
21664 1234897
21703 1234897
21738 1234897
21788 1234897
21826 1234897
21869 1234897
21900 1234897
21923 1234897
21967 1234897
22002 1234897
22070 1234897
22135 1234897
22168 1234897
22198 1234897
22238 1234897
22281 1234897
22311 1234897
22348 1234897
22400 1234897
22440 1234897
22478 1234897
22507 1234897
22532 1234897
22572 1234897
22617 1234897
22662 1234897
22692 1234897
22740 1234897
22785 1234897
22829 1234897
22886 1234897
22925 1234897
22970 1234897
23025 1234897
23098 1234897
23130 1234897
23195 1234897
23235 1234897
23260 1234897
23310 1234897
23341 1234897
23368 1234897
23411 1234897
23475 1234897
23547 1234897
23608 1234897
23688 1234897
23735 1234897
23778 1234897
23810 1234897
23878 1234897
23924 1234897
23952 1234897
24024 1234897
24097 1234897
24151 1234897
24209 1234897
24259 1234897
24312 1234897
24360 1234897
24441 1234897
24473 1234897
24550 1234897
24597 1234897
24657 1234897
24691 1234897
24727 1234897
24803 1234897
24851 1234897
24890 1234897
24926 1234897
24976 1234897
25017 1234897
25043 1234897
25086 1234897
25132 1234897
25196 1234897
25253 1234897
25305 1234897
25333 1234897
25359 1234897
25392 1234897
25426 1234897
25515 1234897
25554 1234897
25624 1234897
25660 1234897
25711 1234897
25767 1234897
25803 1234897
25849 1234897
25881 1234897
25903 1234897
25936 1234897
25970 1234897
26032 1234897
26064 1234897
26096 1234897
26154 1234897
26179 1234897
26210 1234897
26259 1234897
26289 1234897
26340 1234897
26387 1234897
26452 1234897
26511 1234897
26588 1234897
26641 1234897
26692 1234897
26717 1234897
26756 1234897
26803 1234897
26868 1234897
26922 1234897
26949 1234897
26983 1234897
27018 1234897
27064 1234897
27111 1234897
27146 1234897
27198 1234897
27238 1234897
27273 1234897
27347 1234897
27388 1234897
27451 1234897
27506 1234897
27570 1234897
27596 1234897
27634 1234897
27662 1234897
27686 1234897
27723 1234897
27764 1234897
27795 1234897
27839 1234897
27910 1234897
27937 1234897
27986 1234897
28028 1234897
28057 1234897
28087 1234897
28121 1234897
28190 1234897
28217 1234897
28232 1234897
28265 1234897
28317 1234897
28395 1234897
28430 1234897
28488 1234897
28538 1234897
28561 1234897
28601 1234897
28654 1234897
28689 1234897
28749 1234897
28803 1234897
28861 1234897
28890 1234897
28930 1234897
28986 1234897
29020 1234897
29050 1234897
29090 1234897
29137 1234897
29201 1234897
29228 1234897
29300 1234897
29338 1234897
29392 1234897
29449 1234897
29497 1234897
29534 1234897
29579 1234897
29611 1234897
29677 1234897
29724 1234897
29750 1234897
29799 1234897
29856 1234897
29896 1234897
29924 1234897
29983 1234897
30026 1234897
30052 1234897
30110 1234897
30153 1234897
30190 1234897
30232 1234897
30314 1234897
30343 1234897
30382 1234897
30450 1234897
30495 1234897
30533 1234897
30601 1234897
30638 1234897
30702 1234897
30731 1234897
30759 1234897
30816 1234897
30860 1234897
30904 1234897
30952 1234897
30994 1234897
31035 1234897
31098 1234897
31129 1234897
31167 1234897
31213 1234897
31266 1234897
31301 1234897
31339 1234897
31382 1234897
31427 1234897
31489 1234897
31541 1234897
31558 1234897
31593 1234897
31647 1234897
31673 1234897
31713 1234897
31761 1234897
31798 1234897
31841 1234897
31872 1234897
31937 1234897
31996 1234897
32025 1234897
32062 1234897
32100 1234897
32146 1234897
32201 1234897
32246 1234897
32311 1234897
32372 1234897
32413 1234897
32452 1234897
32486 1234897
32516 1234897
32562 1234897
32627 1234897
32684 1234897
32726 1234897
32756 1234897
32779 1234897
32810 1234897
32871 1234897
32932 1234897
32976 1234897
33028 1234897
33062 1234897
33126 1234897
33162 1234897
33210 1234897
33265 1234897
33309 1234897
33358 1234897
33400 1234897
33454 1234897
33509 1234897
33571 1234897
33631 1234897
33677 1234897
33712 1234897
33753 1234897
33796 1234897
33826 1234897
33870 1234897
33909 1234897
33956 1234897
34007 1234897
34047 1234897
34105 1234897
34165 1234897
34208 1234897
34242 1234897
34345 1234897
34374 1234897
34441 1234897
34477 1234897
34514 1234897
34563 1234897
34619 1234897
34650 1234897
34686 1234897
34721 1234897
34776 1234897
34826 1234897
34862 1234897
34914 1234897
34965 1234897
34993 1234897
35043 1234897
35092 1234897
35144 1234897
35203 1234897
35233 1234897
35278 1234897
35307 1234897
35356 1234897
35401 1234897
35452 1234897
35505 1234897
35548 1234897
35589 1234897
35637 1234897
35694 1234897
35735 1234897
35780 1234897
35840 1234897
35883 1234897
35934 1234897
35978 1234897
36016 1234897
36048 1234897
36092 1234897
36154 1234897
36203 1234897
36266 1234897
36314 1234897
36370 1234897
36427 1234897
36490 1234897
36518 1234897
36560 1234897
36589 1234897
36640 1234897
36691 1234897
36746 1234897
36805 1234897
36848 1234897
36875 1234897
36902 1234897
36949 1234897
36989 1234897
37049 1234897
37086 1234897
37129 1234897
37194 1234897
37235 1234897
37299 1234897
37346 1234897
37403 1234897
37429 1234897
37477 1234897
37520 1234897
37554 1234897
37599 1234897
37632 1234897
37679 1234897
37775 1234897
37838 1234897
37876 1234897
37937 1234897
37972 1234897
38023 1234897
38088 1234897
38126 1234897
38172 1234897
38200 1234897
38242 1234897
38304 1234897
38336 1234897
38380 1234897
38415 1234897
38457 1234897
38509 1234897
38545 1234897
38580 1234897
38617 1234897
38672 1234897
38702 1234897
38743 1234897
38790 1234897
38843 1234897
38896 1234897
38936 1234897
38958 1234897
39008 1234897
39064 1234897
39120 1234897
39172 1234897
39223 1234897
39249 1234897
39315 1234897
39349 1234897
39380 1234897
39406 1234897
39447 1234897
39483 1234897
39520 1234897
39559 1234897
39580 1234897
39626 1234897
39664 1234897
39717 1234897
39760 1234897
39810 1234897
39853 1234897
39939 1234897
39979 1234897
40015 1234897
40051 1234897
40104 1234897
40165 1234897
40203 1234897
40256 1234897
40310 1234897
40347 1234897
40390 1234897
40432 1234897
40469 1234897
40505 1234897
40534 1234897
40603 1234897
40648 1234897
40705 1234897
40742 1234897
40791 1234897
40845 1234897
40884 1234897
40922 1234897
40970 1234897
41039 1234897
41102 1234897
41164 1234897
41203 1234897
41223 1234897
41275 1234897
41316 1234897
41340 1234897
41383 1234897
41435 1234897
41490 1234897
41536 1234897
41563 1234897
41622 1234897
41676 1234897
41719 1234897
41766 1234897
41788 1234897
41824 1234897
41873 1234897
41926 1234897
41961 1234897
42019 1234897
42040 1234897
42114 1234897
42150 1234897
42190 1234897
42227 1234897
42261 1234897
42316 1234897
42374 1234897
42419 1234897
42456 1234897
42494 1234897
42542 1234897
42577 1234897
42633 1234897
42665 1234897
42710 1234897
42732 1234897
42771 1234897
42813 1234897
42857 1234897
42879 1234897
42921 1234897
42959 1234897
42993 1234897
43028 1234897
43079 1234897
43137 1234897
43171 1234897
43234 1234897
43275 1234897
43316 1234897
43353 1234897
43417 1234897
43453 1234897
43520 1234897
43547 1234897
43628 1234897
43651 1234897
43694 1234897
43746 1234897
43763 1234897
43803 1234897
43852 1234897
43895 1234897
43946 1234897
44001 1234897
44033 1234897
44097 1234897
44140 1234897
44184 1234897
44237 1234897
44283 1234897
44333 1234897
44365 1234897
44416 1234897
44458 1234897
44486 1234897
44529 1234897
44567 1234897
44600 1234897
44624 1234897
44658 1234897
44718 1234897
44757 1234897
44809 1234897
44848 1234897
44909 1234897
44969 1234897
};
\end{axis}

\begin{axis}[
axis y line=right,
axis x line*=bottom,
height=\figureheight,
tick align=outside,
width=\figurewidth,
x grid style={darkgrey176},
xmin=-2242.15, xmax=47217.15,
xtick pos=left,
xtick style={color=black},
y grid style={darkgrey176},
ylabel={Quality (PSNR, dB)},
ymin=18.2211750984192, ymax=25.4101553916931,
ytick pos=right,
ytick style={color=black},
yticklabel style={anchor=west},
every axis y label/.append style ={darkorange25512714},
]
\addplot [semithick, darkorange25512714]
table {%
100 18.5479469299316
200 19.0962600708008
300 19.3471508026123
400 19.6698150634766
500 19.81960105896
600 19.9756565093994
700 20.06764793396
800 20.2530078887939
900 20.3486404418945
1000 20.4430904388428
1100 20.6239967346191
1200 20.722806930542
1300 20.7396411895752
1400 20.8494873046875
1500 20.8649978637695
1600 20.9626579284668
1700 21.0628356933594
1800 21.1485748291016
1900 21.1456489562988
2000 21.1926422119141
2100 21.2265605926514
2200 21.2683219909668
2300 21.4067058563232
2400 21.3638134002686
2500 21.4143390655518
2600 21.429988861084
2700 21.58642578125
2800 21.7610321044922
2900 21.6167812347412
3000 21.7171058654785
3100 21.409839630127
3200 21.6420764923096
3300 21.5991153717041
3400 21.6350326538086
3500 21.8617401123047
3600 21.956184387207
3700 22.0800285339355
3800 22.0525074005127
3900 22.0393295288086
4000 22.2152671813965
4100 22.2354106903076
4200 22.2135028839111
4300 22.2782249450684
4400 22.2446823120117
4500 22.3455543518066
4600 22.4311256408691
4700 22.6276321411133
4800 22.4106140136719
4900 22.5678272247314
5000 22.5888652801514
5100 22.6992816925049
5200 22.8294124603271
5300 22.8263339996338
5400 22.7323398590088
5500 22.8719081878662
5600 22.9329299926758
5700 23.0724601745605
5800 22.9091110229492
5900 23.1213893890381
6000 23.0110130310059
6100 22.5239715576172
6200 22.8342475891113
6300 22.9847202301025
6400 22.9214191436768
6500 23.0886268615723
6600 23.1471920013428
6700 23.2455520629883
6800 23.3690452575684
6900 23.3329963684082
7000 23.4297504425049
7100 23.4177703857422
7200 23.6476860046387
7300 23.5499286651611
7400 23.6969928741455
7500 23.613130569458
7600 23.69069480896
7700 23.5822925567627
7800 23.5314826965332
7900 23.7713413238525
8000 23.902551651001
8100 23.7175426483154
8200 23.880542755127
8300 23.8481521606445
8400 23.955472946167
8500 23.9714622497559
8600 23.9382209777832
8700 23.9481315612793
8800 24.0522079467773
8900 24.0245628356934
9000 24.0978012084961
9100 23.2941436767578
9200 23.6969432830811
9300 23.7470779418945
9400 23.7936782836914
9500 23.9986133575439
9600 24.0018978118896
9700 24.0213756561279
9800 23.9551010131836
9900 24.2688102722168
10000 24.174934387207
10100 24.1977024078369
10200 24.2604675292969
10300 24.3829536437988
10400 24.3433685302734
10500 24.2784614562988
10600 24.325174331665
10700 24.2495212554932
10800 24.2823581695557
10900 24.3344440460205
11000 24.4436302185059
11100 24.3526477813721
11200 24.5203628540039
11300 24.4175224304199
11400 24.3741874694824
11500 24.4395942687988
11600 24.5580062866211
11700 24.554048538208
11800 24.4785995483398
11900 24.5914363861084
12000 24.5716247558594
12100 23.6112880706787
12200 24.2756042480469
12300 24.3024215698242
12400 24.4510364532471
12500 24.4226245880127
12600 24.5158252716064
12700 24.588888168335
12800 24.5625419616699
12900 24.4635848999023
13000 24.601167678833
13100 24.6359615325928
13200 24.5743999481201
13300 24.6587524414062
13400 24.6231899261475
13500 24.744291305542
13600 24.7606372833252
13700 24.6085891723633
13800 24.6630744934082
13900 24.7634258270264
14000 24.6651191711426
14100 24.7041645050049
14200 24.6622085571289
14300 24.7420864105225
14400 24.7857303619385
14500 24.7678699493408
14600 24.6955795288086
14700 24.7481918334961
14800 24.8641033172607
14900 24.7832489013672
15000 24.8244132995605
15100 24.4996376037598
15200 24.3760795593262
15300 24.5861396789551
15400 24.566987991333
15500 24.743953704834
15600 24.7110729217529
15700 24.5946750640869
15800 24.7549495697021
15900 24.7611923217773
16000 24.8127040863037
16100 24.8351287841797
16200 24.7191944122314
16300 24.638011932373
16400 24.7428894042969
16500 24.7971897125244
16600 24.8906917572021
16700 24.8294811248779
16800 24.7615833282471
16900 24.8885135650635
17000 24.8575248718262
17100 24.9007949829102
17200 24.923131942749
17300 24.8719501495361
17400 24.8122749328613
17500 24.7666435241699
17600 24.850004196167
17700 24.84157371521
17800 24.9143619537354
17900 24.8616828918457
18000 24.9055233001709
18100 24.9833068847656
18200 24.8816242218018
18300 24.840482711792
18400 24.9039440155029
18500 24.8166561126709
18600 24.9632911682129
18700 24.9386177062988
18800 24.9358234405518
18900 24.8611030578613
19000 24.9302310943604
19100 24.9551849365234
19200 24.9422760009766
19300 24.9838123321533
19400 24.9658279418945
19500 25.00368309021
19600 24.9363536834717
19700 24.9326782226562
19800 24.9108047485352
19900 24.8909778594971
20000 24.9059410095215
20100 24.4575214385986
20200 24.4158306121826
20300 24.5599060058594
20400 24.598949432373
20500 24.5708713531494
20600 24.6198539733887
20700 24.6256046295166
20800 24.6252174377441
20900 24.6560897827148
21000 24.7229766845703
21100 24.6485042572021
21200 24.7049007415771
21300 24.7177429199219
21400 24.7223777770996
21500 24.7511672973633
21600 24.653076171875
21700 24.7334136962891
21800 24.7062797546387
21900 24.7301845550537
22000 24.7422618865967
22100 24.7823066711426
22200 24.7534465789795
22300 24.7360572814941
22400 24.7138061523438
22500 24.7613868713379
22600 24.7479763031006
22700 24.7803573608398
22800 24.7995758056641
22900 24.7810363769531
23000 24.8008632659912
23100 24.7775955200195
23200 24.8197784423828
23300 24.7801761627197
23400 24.758861541748
23500 24.7966861724854
23600 24.8204822540283
23700 24.7765636444092
23800 24.7673072814941
23900 24.7856884002686
24000 24.7976875305176
24100 24.7712249755859
24200 24.8011035919189
24300 24.8019638061523
24400 24.7854137420654
24500 24.8317012786865
24600 24.8779945373535
24700 24.8174839019775
24800 24.8319931030273
24900 24.8354454040527
25000 24.8247470855713
25100 24.8091297149658
25200 24.7991333007812
25300 24.8373603820801
25400 24.7698173522949
25500 24.8328056335449
25600 24.8023796081543
25700 24.8715000152588
25800 24.8208084106445
25900 24.801643371582
26000 24.8153381347656
26100 24.8914909362793
26200 24.8264923095703
26300 24.7581996917725
26400 24.8200836181641
26500 24.85524559021
26600 24.8197803497314
26700 24.8289756774902
26800 24.802267074585
26900 24.8689308166504
27000 24.8541774749756
27100 24.8524742126465
27200 24.8272075653076
27300 24.8391990661621
27400 24.8556385040283
27500 24.8323669433594
27600 24.8817882537842
27700 24.8591270446777
27800 24.8602313995361
27900 24.85888671875
28000 24.8243389129639
28100 24.8331031799316
28200 24.8339061737061
28300 24.858268737793
28400 24.8895988464355
28500 24.8902397155762
28600 24.8733043670654
28700 24.9032230377197
28800 24.8733291625977
28900 24.8958683013916
29000 24.8750057220459
29100 24.8832511901855
29200 24.8502101898193
29300 24.8304004669189
29400 24.8263187408447
29500 24.9067497253418
29600 24.8863925933838
29700 24.8718280792236
29800 24.9082851409912
29900 24.9136695861816
30000 24.8680629730225
30100 24.8374404907227
30200 24.8948211669922
30300 24.8701305389404
30400 24.9219169616699
30500 24.889965057373
30600 24.8507690429688
30700 24.889892578125
30800 24.9205455780029
30900 24.892448425293
31000 24.8681907653809
31100 24.8488731384277
31200 24.8856410980225
31300 24.9064903259277
31400 24.8940238952637
31500 24.8444595336914
31600 24.9005584716797
31700 24.8957252502441
31800 24.8872814178467
31900 24.9151401519775
32000 24.9099941253662
32100 24.8959045410156
32200 24.9453907012939
32300 24.8872814178467
32400 24.88551902771
32500 24.8768100738525
32600 24.8852939605713
32700 24.8641166687012
32800 24.8928165435791
32900 24.896900177002
33000 24.8514041900635
33100 24.9213085174561
33200 24.9042110443115
33300 24.9174976348877
33400 24.8832092285156
33500 24.9340324401855
33600 24.8656387329102
33700 24.919849395752
33800 24.8673248291016
33900 24.8975009918213
34000 24.9354801177979
34100 24.8745193481445
34200 24.8673191070557
34300 24.8868942260742
34400 24.9546203613281
34500 24.8977375030518
34600 24.9067401885986
34700 24.8581695556641
34800 24.8762512207031
34900 24.8795337677002
35000 24.9229793548584
35100 24.8556346893311
35200 24.9314060211182
35300 24.9169025421143
35400 24.8926372528076
35500 24.9221363067627
35600 24.8693027496338
35700 24.9560298919678
35800 24.8965969085693
35900 24.9008331298828
36000 24.968168258667
36100 24.8926124572754
36200 24.8710613250732
36300 24.9597854614258
36400 24.8815975189209
36500 24.8976497650146
36600 24.9152393341064
36700 24.9328880310059
36800 24.8772258758545
36900 24.8575839996338
37000 24.9587936401367
37100 24.9623947143555
37200 24.9155101776123
37300 24.9071807861328
37400 24.9520454406738
37500 24.9206771850586
37600 24.919584274292
37700 24.9121322631836
37800 24.9123210906982
37900 24.8602905273438
38000 24.9531269073486
38100 24.9438877105713
38200 24.9040374755859
38300 24.9201278686523
38400 24.8716659545898
38500 24.929027557373
38600 24.8728141784668
38700 24.9457321166992
38800 24.9114284515381
38900 24.9169292449951
39000 24.8578624725342
39100 24.8979797363281
39200 24.9028701782227
39300 24.8704395294189
39400 24.9025344848633
39500 24.9186496734619
39600 24.9388103485107
39700 24.8909606933594
39800 24.9278717041016
39900 24.9420623779297
40000 24.9228534698486
40100 24.9245166778564
40200 24.9320106506348
40300 24.9256210327148
40400 24.8635807037354
40500 24.8955478668213
40600 24.903299331665
40700 24.9149398803711
40800 24.9135589599609
40900 24.9324836730957
41000 24.9202041625977
41100 24.9363803863525
41200 24.9606590270996
41300 24.9285831451416
41400 24.9148483276367
41500 24.8917465209961
41600 24.9328098297119
41700 24.9016647338867
41800 24.9430446624756
41900 24.9365367889404
42000 24.9077892303467
42100 24.9466419219971
42200 24.9295425415039
42300 24.9106159210205
42400 24.9108409881592
42500 24.8902702331543
42600 24.9491901397705
42700 24.9318447113037
42800 24.9394721984863
42900 24.9010162353516
43000 24.8926124572754
43100 24.9618949890137
43200 24.9987640380859
43300 25.0464782714844
43400 25.0159511566162
43500 24.9744625091553
43600 25.0394191741943
43700 25.0142765045166
43800 25.0060253143311
43900 25.0302028656006
44000 24.9471321105957
44100 25.0431652069092
44200 25.0489311218262
44300 25.0710430145264
44400 25.0645065307617
44500 25.0202579498291
44600 24.9859619140625
44700 25.0833835601807
44800 24.978609085083
44900 25.0564651489258
};
\end{axis}

\end{tikzpicture}
  \caption{Overview of the training process consisting of four phases. During {\bf warm-up}, the model learns the 3D information by increasing the number of Gaussian splats. The {\bf pruning} stage reduces the number of Gaussians while maintaining reconstruction performance. Density control is applied only until the end of the pruning phase. In the {\bf vector quantization} phase, reconstruction quality initially degrades but recovers after sufficient training. Finally, the {\bf fine-tuning} removes constraints imposed by noise substitution, further refining the final results.\looseness-1}
  \label{fig:training}
\end{figure}

\paragraph{Training process}
The entire process comprises four phases as illustrated in \cref{fig:training}: the warm-up, pruning, vector quantization, and fine-tuning.

Consider a set of $N$ Gaussians splats, denoted as $\{G_i\}_{i=1}^N$. Each splat $G$ contains features $G = (\vx, o, \vs, \vr, \vc, \vc^{sh})$ as described in \cref{subsec:bg_gs}. The training process begins with a \textbf{warm-up phase}, spanning the first 15k iterations. During this phase, the training procedure aligns precisely with the standard 3DGS \cite{kerbl20233d_3dgs}, including the adaptive density control for splat densification and pruning. 

The \textbf{pruning phase} occurs between 15k and 20k iterations, during which Gaussian splats are further pruned using opacity regularization, as introduced in CompGS \cite{navaneet2024compgs}. In this phase, the training objective incorporates an additional regularization loss term, defined as $\mathcal{L}_{opacity} = \sum_{i=1}^N o_i$. Splats with low opacity are subsequently removed to enhance the overall compaction.

The \textbf{vector quantization phase} begins after the initial 20k iterations. In this work, we quantize all parameters except for the coordinates $\vx$ and opacity $o$, as quantizing these parameters would result in obvious quality degradation. We construct codebooks $\mathcal{C}_{s} \in \mathbb{R}^{2^{K_s} \times 3}, \mathcal{C}_{r}  \in \mathbb{R}^{2^{K_r} \times 4}, \mathcal{C}_{c}  \in \mathbb{R}^{2^{K_c} \times 3}, \mathcal{C}_{sh}  \in \mathbb{R}^{2^{K_{sh}} \times 45}$ for four attributes associated with the covariance matrix and colours, where $K_s, K_r, K_c, K_{sh} \in \mathbb{N}$ denote the respective codebook bitrates. The codebook entries are initialized as the centroids of clustered feature distributions by K-means. The quantized Gaussians splats are then represented as: $
\tilde{G}_q = (\vx, o, \tilde{\vs}_q, \tilde{\vr}_q, \tilde{\vc}_q, \tilde{\vc}^{sh}_q),$
where $ (\tilde{\vs}_q, \tilde{\vr}_q, \tilde{\vc}_q, \tilde{\vc}^{sh}_q)$ are quantized features obtained from codebooks $\mathcal{C}_{s}, \mathcal{C}_{r}, \mathcal{C}_{c}$, $\mathcal{C}_{sh}$ using \cref{eq:nsvq}. During this phase, the model utilises $\tilde{G}_q$ for $\alpha$-blending as described in \cref{eq:alpha-blending}. Both model parameters and all codebooks are trained jointly, with periodic replacement of codebook entries to increase the codebook entries usage. \sloppy

In the final 2k iterations, the model undergoes a \textbf{fine-tuning phase} with frozen code assignment, which means the code indices $k_{\vx}^*$ in \cref{eq:hard_quantization} are fixed, without the need for a nearest code search. Meanwhile, the noise-substitution in quantized features is skipped, meaning the model is trained with hard-quantized features $G_q = (\vx, o, \vs_q, \vr_q, \vc_q, \vc^{sh}_q)$.

In the end, the model is stored as $(\{\hat{G}_i\}_{i=1}^N, \{\mathcal{C}_{i}\}_{i\in \{s, r, c, sh\}})$, where the quantized attributes are replaced by their corresponding codebook entries: $\hat{G}=(\vx, o,$ $ k_{\vs}^*, k_{\vr}^*, k_{\vc}^*,k_{\vc^{sh}}^*)$ which represent the final quantized Gaussian splats. Each index, ranging from $1$ to $2^K$, requires $K$ bits. Consequently, all indices are stored in a compact bitstream format within a binary file. \sloppy

\begin{table}[t!]
  \small\centering
  \caption{Benchmark comparison with baseline methods. Values are bold as the best results among all SP-based methods. (Baselines are collected from the benchmark  \cite{bagdasarian20243dgs_3dgszip}.) The model size unit is MB.
  }
  \label{table:benchmark}
  \setlength{\tabcolsep}{5pt}
  \newcommand{\nan}{---}
  \newcommand{\boldvalue}[1]{\textbf{#1}}
  \newcommand{\rotate}[2]{\parbox[c]{2mm}{\multirow{#1}{*}{\tikz[baseline]\node[anchor=base,rotate=90,font=\tiny,align=center]{#2};}}}
  \resizebox{\textwidth}{!}{%
\begin{tabular}{ll|cccc|cccc|cccc}
\toprule
 && \multicolumn{4}{c}{Mip-NeRF 360} & \multicolumn{4}{c}{Tanks and Temples} & \multicolumn{4}{c}{Deep Blending}
\\ 
   &Methods & PSNR$\uparrow$ & SSIM$\uparrow$ & LPIPS$\downarrow$ & Size$\downarrow$ &PSNR$\uparrow$ & SSIM$\uparrow$ & LPIPS$\downarrow$ & Size$\downarrow$ & PSNR$\uparrow$ & SSIM$\uparrow$ & LPIPS$\downarrow$ & Size$\downarrow$
\\\midrule
\rowcolor{blue!10} &\modelname\ (16k) (ours) & \boldvalue{27.28} & 0.807 & 0.239 & \boldvalue{16.38} & \boldvalue{23.62} & \boldvalue{0.842} & 0.190 & \boldvalue{11.02} & \boldvalue{29.90} & \boldvalue{0.906} & \boldvalue{0.249} & \boldvalue{11.49} \\
&CompGS (16k) & 27.03 & 0.804 & 0.243 & 18 & 23.39 & 0.836 & 0.200 & 12 & \boldvalue{29.90} & \boldvalue{0.906} & 0.252 & 12 \\
&Reduced3DGS & 27.10 & \boldvalue{0.809} & \boldvalue{0.226} & 29 & 23.57 & 0.840 & \boldvalue{0.188} & 14 & 29.63 & 0.902 & \boldvalue{0.249} & 18 \\
&Compact3DGS & 27.08 & 0.798 & 0.247 & 48.8 & 23.32 & 0.831 & 0.201 & 39.4 & 29.79 & 0.901 & 0.258 & 43.2 \\
&LightGaussians & \boldvalue{27.28} & 0.805 & 0.243 & 42 & 23.11 & 0.817 & 0.231 & 22 & \nan & \nan & \nan & \nan \\
\rotate{-6}{SP-based}\cellcolor{white}&Compressed3D & 26.98 & 0.801 & 0.238 & 28.8 & 23.32 & 0.832 & 0.194 & 17.28 & 29.38 & 0.898 & 0.253 & 25.3 \\
\midrule
\rotate{3}{ML-based} &HAC (lowrate) & 27.53 & 0.807 & 0.238 & 15.26 & 24.04 & 0.846 & 0.187 & 8.1 & 29.98 & 0.902 & 0.269 & 4.35 \\
&SOG & 27.08 & 0.799 & 0.230 & 38.42 & 23.56 & 0.837 & 0.186 & 21.72 & 29.26 & 0.894 & 0.268 & 16.92 \\
&ContextGS (lowrate) & 27.62 & 0.808 & 0.237 & 12.68 & 24.12 & 0.849 & 0.186 & 9.443 & 30.09 & 0.907 & 0.265 & 3.485 \\
\midrule
 &3DGS & 27.21 & 0.815 & 0.214 & 734 & 23.14 & 0.841 & 0.183 & 411 & 29.41 & 0.903 & 0.243 & 676 \\
\bottomrule
\end{tabular} %
}
\end{table}

\begin{figure}[t!]
  \centering\small
  \setlength{\figurewidth}{.6\textwidth}
  \setlength{\figureheight}{.6\textwidth}
  \pgfplotsset{
    y tick label style={rotate=90},
    semithick/.append style={line width=3pt,opacity=.8},
    legend style={font=\footnotesize},
    ylabel style={font=\large},
    xlabel style={font=\large},
    grid style={dashed,lightgray},
    legend style={draw=none,inner xsep=2pt, inner ysep=0.5pt, nodes={inner sep=1.5pt, text depth=0.1em},fill=white,fill opacity=0.8}
    }
  \foreach \dataset/\nicename [count=\j] in {MipNeRF360/{Mip-NeRF 360}, TanksAndTemples/{Tanks and Temples}, DeepBlending/{Deep Blending}} {
    \begin{subfigure}[b]{0.02\textwidth}
      \centering 
      \tikz\node[rotate=90,text width=.62\figureheight,align=center,font=\bf\strut,scale=.8]{\nicename};
    \end{subfigure}%
    \foreach \metric [count=\i] in {PSNR, SSIM, LPIPS} {
      \begin{subfigure}[b]{0.31\textwidth}
        \raggedleft 
        \scalebox{.5}{\input{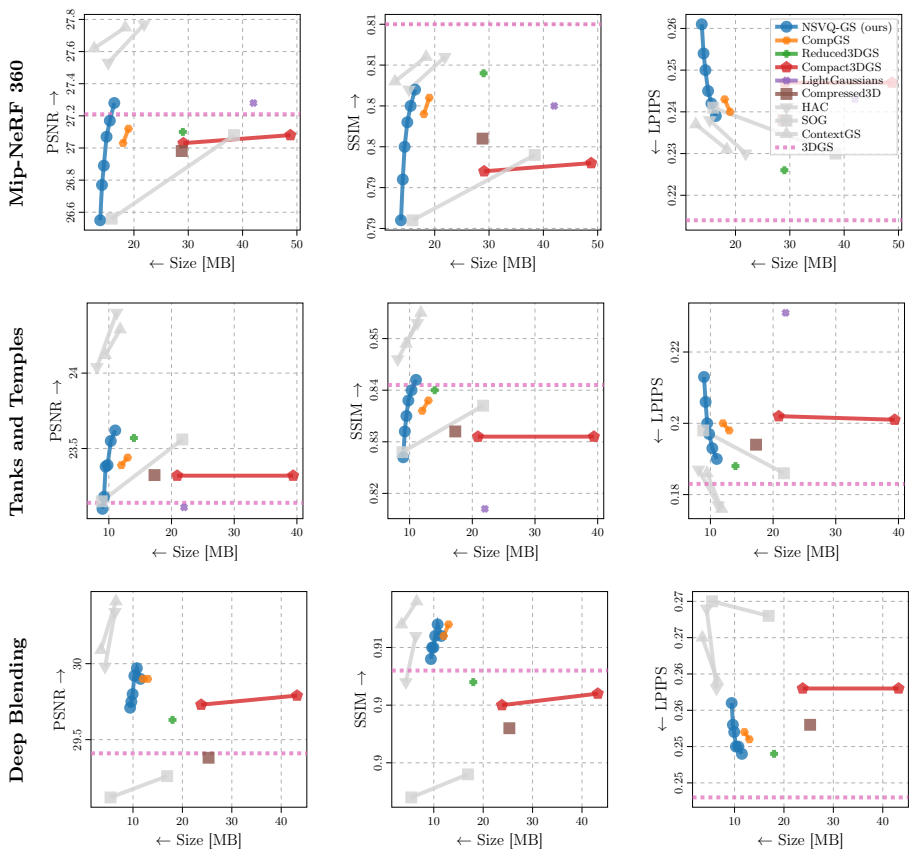}}
      \end{subfigure}
  }}
  \vspace*{-1.5em}
  \definecolor{steelblue31119180}{RGB}{31,119,180}
  \definecolor{lightgrey}{RGB}{211,211,211}
  \caption{The comparison of reconstruction quality across baselines. In each sub-figure, the $x$-axis denotes the model size in Megabytes, $y$-axis denotes the metrics on the reconstruction result. Each method comprising different sub-methods is plotted as multiple connected points. Our \textcolor{steelblue31119180}{\textbf{\modelname}}\ performs best within the category of SP-based methods, whereas ML-based methods (in \textcolor{lightgrey}{\bf gray}) can boost performance further while losing some of the appealing 3DGS propeties.   \looseness-1
  }
  \label{fig:quantitative_scatter}
\end{figure} 
\section{Experiments}
\label{sec:experiments}
We evaluate \modelname\ in 3DGS compression, aiming to demonstrate two key aspects: compression efficiency---how well we reduce storage costs while maintaining model fidelity, and rendering performance---how the compressed models perform in reconstruction. We follow the benchmarking protocols established in 3DGS.zip \cite{bagdasarian20243dgs_3dgszip}, with main comparisons to the closest prior work, CompGS~\cite{navaneet2024compgs}.

\subsection{Settings}
\label{subsec:ex_settings}
\paragraph{Data sets}
We evaluate our method for real-world 3D scene reconstruction tasks on the standard benchmark data sets, following the conventions established in 3DGS \cite{kerbl20233d_3dgs}. The benchmark consists of three data sets: Mip-NeRF360~\cite{ds_Mip-NeRF} (9 scenes), Tanks \& Temples~\cite{ds_tanks_and_temples} (2 scenes), and Deep Blending~\cite{ds_deep_blending} (2 scenes). These data sets cover a diverse range of real-world scenarios, including both unbounded outdoor environments and complex indoor settings. Train and test data split adheres to the methodology suggested by Mip-NeRF360~\cite{ds_Mip-NeRF}, where the test set comprises every 8\textsuperscript{th} image (\ie, images with indices satisfying $i \mod 8 \equiv 0$), while the remaining images are allocated as the training set.

\paragraph{Implementation}
Our training process consists of four phases: (1)~warm-up phase for the first 15k iterations, (2)~pruning phase during 15k--20k iterations, (3) vector quantization phase where  Gaussian splats are trained jointly with codebooks during 20k--43k iterations,  and (4)~the fine-tuning phase with frozen codebook assignment during the 43k--45k iterations. In all experiments, we keep the convention established in CompGS \cite{navaneet2024compgs}, by setting the bitrates of codebook $K_s = K_r = 4K_c = 4K_{sh}$. The codebook bitrates notation follows a power-of-two scaling, where `16k' means $K_s = 2^{14} = 16384$, `8k' means $K_s = 2^{13} = 8192$, and so on.
In practice, all experiments are conducted on a computing server equipped with a Nvidia A100 GPU and 32~GB memory. Our programming implementation is based on the pytorch-1.12.0 package and several submodules from the 3DGS codebase \cite{kerbl20233d_3dgs}. On average, training one scene with `4k' bitrate settings requires around 100 minutes.

\paragraph{Metrics}
To evaluate the performance of NVS, we utilize widely recognized metrics. The Peak Signal-to-Noise Ratio (PSNR) quantifies the ratio between the maximum possible power of signals (\ie, the ground truth image) and the power of corrupting noise. The Structural Similarity Index Measure (SSIM) evaluates perceptive quality by accounting for luminance, contrast, and structure degradation in synthetic images. The Learned Perceptual Image Patch Similarity (LPIPS) computes image similarity using a pre-defined NN designed to align with human perception \cite{Zhang_2018_CVPR_lpips}. Beyond image-based reconstruction quality, we also report model size (in megabytes) as a metric to evaluate the compression efficiency, reflecting the representation compactness.

\paragraph{Baselines}
We compare our methods with baseline methods, including original 3DGS~(30k)~\cite{kerbl20233d_3dgs}, five SP-based methods: CompGS~\cite{navaneet2024compgs}, Reduced3DGS~\cite{10.1145/3651282_reduced3dgs}, Compact3DGS \cite{Lee_2024_CVPR_compact3dgs}, LightGaussians~\cite{fan2023lightgaussian} and Compressed3D~\cite{Niedermayr_2024_CVPR_compressed3dgs}, and three additional ML-based methods: HAC~\cite{chen2025hac}, SOG~\cite{morgenstern2024compact_SOG}, and ContextGS~\cite{wang2024contextgs}.

\subsection{Results}
\paragraph{Quantitative results}
We evaluate the reconstruction performance using four quantitative metrics summarized in \cref{table:benchmark} and visualized in \cref{fig:quantitative_scatter}. 

In \cref{table:benchmark}, the methods are categorized into three clusters: SP-based methods, ML-based methods and original 3DGS~(30k). The results demonstrate that our \modelname~(16k) reaches an optimal balance between compression efficiency and reconstruction quality across all SP-based baselines. Compared to 3DGS~(30k), our \modelname attains higher PSNRs across all data sets while utilizing only $2.2\%$ of the storage consumption on average. 

However, it is important to note that the table presents only one sub-method for each method, whereas each model may encompass multiple sub-methods, reflecting varying trade-offs between compression ratio and reconstruction quality. To facilitate a more comprehensive visualization, the scatter plots in \cref{fig:quantitative_scatter} include all sub-methods. Specifically, the sub-methods of our \modelname differ in codebook bitrates, ranging from `0.5k' to `16k'. This comparison demonstrates that our model outperforms all other SP-based GS compression baselines, particularly the best-performing SP-based baseline, CompGS. It is observed that reducing bitrates leads to a degradation in reconstruction with decreasing storage benefits, as the primary storage consumption is attributed to non-quantized features. Therefore, to achieve a better compresion model, it is advisable to retain relative high codebook bitrates and focusing on optimizing non-quantized features. However, the SSIM and LPIPS of \modelname are generally worse than the PSNR compared to 3DGS on all datasets, possibly due to the lack of locality prior and global sense in our compresion in our technique.

\begin{figure*}[t]
  \raggedleft
  \setlength{\figurewidth}{0.23\textwidth}
  \setlength{\figureheight}{\figurewidth}
  \foreach \method/\methodname [count=\j] in {gt/{\strut Ground truth},compgs0.5k/{\strut ~CompGS\ (0.5k)~}, newvq0.5k/{\strut \modelname\ (0.5k)}} {
  \begin{tikzpicture}[
      image/.style = {inner sep=0pt, outer sep=0pt, 
        minimum width=.95\figurewidth, anchor=north west, text width=.95\figurewidth}, 
      node distance = 1pt and 1pt, every node/.style={font= {\tiny}}, 
      label/.style = {font={\strut\footnotesize\bf},anchor=south,inner sep=0pt}, 
      spy using outlines={rectangle, magnification=1, size=0.475\figurewidth}
    ] 

    \foreach \scene [count=\i] in {bicycle,bonsai,counter,garden} {
      \node[image] (img-\i) at (\i*\figurewidth,0) 
        {\includegraphics[width=.95\figurewidth]{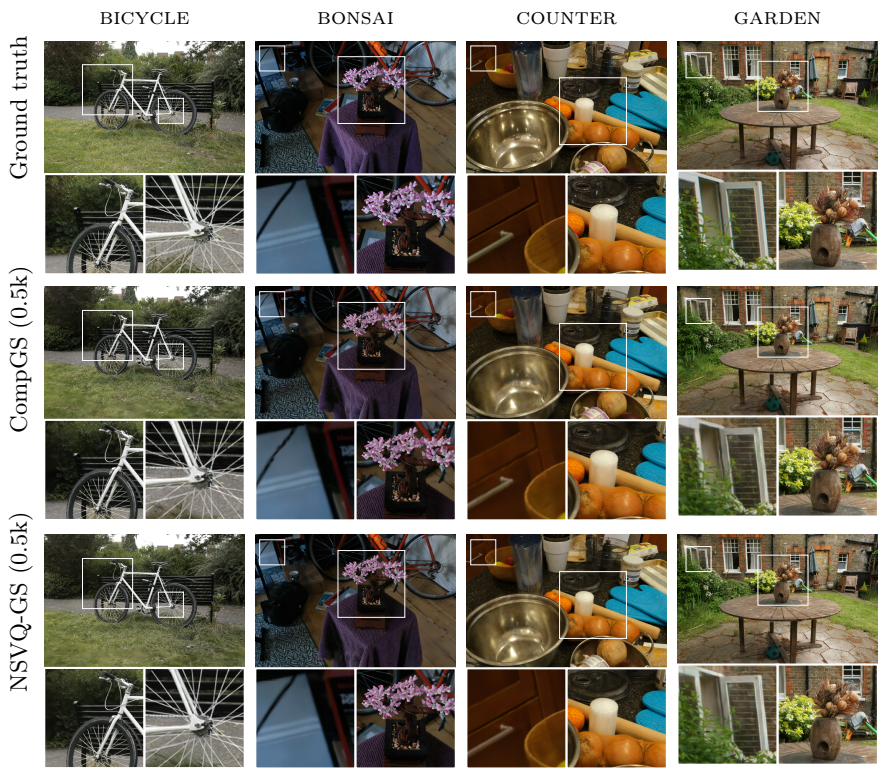}};

      \ifnum\j=1
        \node[anchor=south,font=\strut\sc] at (img-\i.north) {\scene};
      \fi

      \ifnum\i=1
        \node[anchor=south,font=\strut,rotate=90,anchor=south] at (img-\i.west) {\methodname};
      \fi

      \coordinate (spypoint-\i-a) at ($(img-\i.south west)$);
      \coordinate (spypoint-\i-b) at ($(img-\i.south east)$);      
      
    }

    \coordinate (spot-1-a) 
      at ($(img-1.south west) + (0.3\figurewidth,.4\figurewidth)$); %
    \spy[white,magnification=2] on 
      (spot-1-a) in node[anchor=north west] at (spypoint-1-a);
    \coordinate (spot-1-b) 
      at ($(img-1.south west) + (0.6\figurewidth,.3\figurewidth)$); %
    \spy[white,magnification=4] on 
      (spot-1-b) in node[anchor=north east] at (spypoint-1-b);

    \coordinate (spot-2-a) 
      at ($(img-2.south west) + (0.08\figurewidth,.55\figurewidth)$); %
    \spy[white,magnification=4] on 
      (spot-2-a) in node[anchor=north west] at (spypoint-2-a);
    \coordinate (spot-2-b) 
      at ($(img-2.south west) + (0.55\figurewidth,.4\figurewidth)$); %
    \spy[white,magnification=1.5] on 
      (spot-2-b) in node[anchor=north east] at (spypoint-2-b);  

    \coordinate (spot-3-a) 
      at ($(img-3.south west) + (0.08\figurewidth,.55\figurewidth)$); %
    \spy[white,magnification=4] on 
      (spot-3-a) in node[anchor=north west] at (spypoint-3-a);
    \coordinate (spot-3-b) 
      at ($(img-3.south west) + (0.6\figurewidth,.3\figurewidth)$); %
    \spy[white,magnification=1.5] on 
      (spot-3-b) in node[anchor=north east] at (spypoint-3-b);   

    \coordinate (spot-4-a) 
      at ($(img-4.south west) + (0.1\figurewidth,.5\figurewidth)$); %
    \spy[white,magnification=4] on 
      (spot-4-a) in node[anchor=north west] at (spypoint-4-a);
    \coordinate (spot-4-b) 
      at ($(img-4.south west) + (0.5\figurewidth,.4\figurewidth)$); %
    \spy[white,magnification=2] on 
      (spot-4-b) in node[anchor=north east] at (spypoint-4-b);
      
  \end{tikzpicture}~~~\\[-8pt]
  }    
  \vspace*{.05in}
  \caption{
    Qualitative comparison between ground truth, CompGS (0.5k), and \modelname\ (0.5k) (ours). Our \modelname\ captures difficult sharp boundaries and straight lines better compared to CompGS (see, \eg, {\sc bicycle}). This becomes clearer at stronger compression constraints (low codebook bitrates).}
  \label{fig:qualitative_0.5k}
  \vspace*{-.1in}
\end{figure*} 
\begin{figure*}[!t]
  \centering
  \hspace*{0.03\textwidth}
  \foreach \method [count=\i] in {{Ground truth},{3DGS},{CompGS (16k)},{\modelname\ (16k)}}{
    \begin{subfigure}[c]{0.22\textwidth}
      \centering\strut\method
    \end{subfigure}
  }
  \foreach \scene [count=\j] in {flowers, treehill, drjohnson, kitchen, bonsai, train, truck} {
  \begin{tabular}{c}
    \begin{subfigure}[c]{0.03\textwidth}
      \tikz\node[rotate=90,font=\sc\strut]{\scene};
    \end{subfigure}
    \foreach \method [count=\i] in {gt,3dgs,compgs16k,newvq16k}{
      \begin{subfigure}[c]{0.22\textwidth} %
        \centering
        \includegraphics[width=\textwidth]{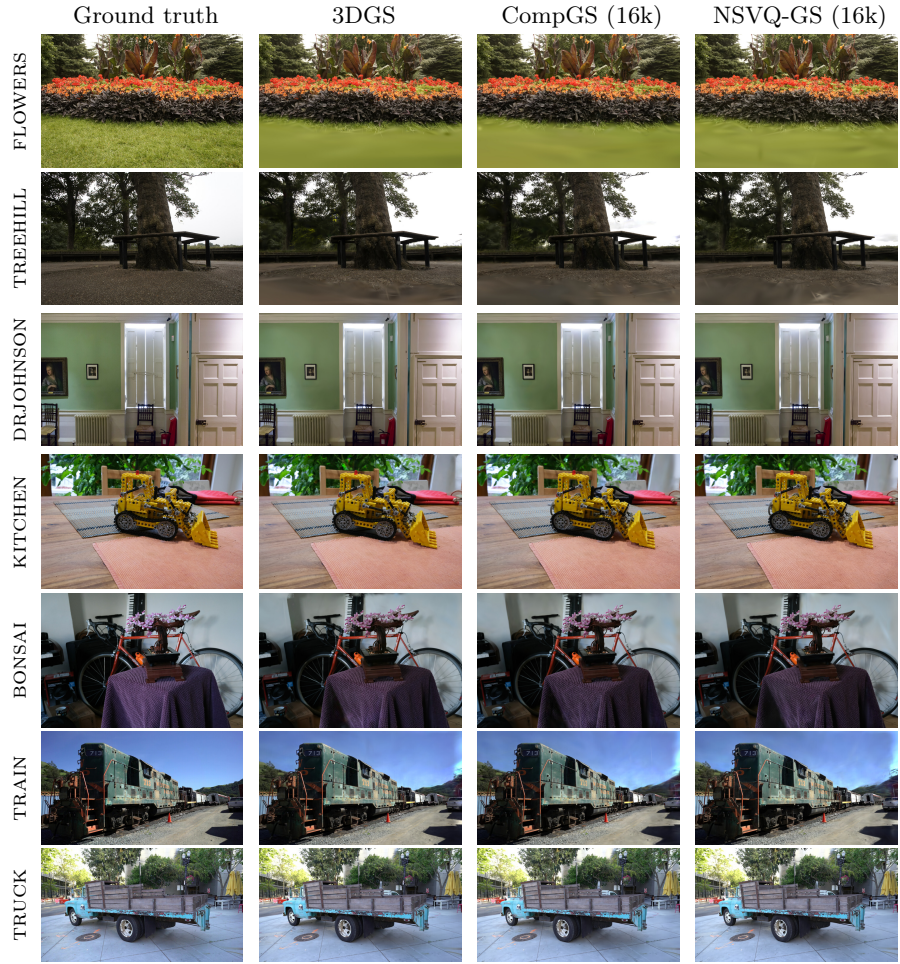}
      \end{subfigure}
      \vspace*{-1pt}
    }
  \end{tabular}\\[.5em]
  }
  \vspace*{-.1in}
  \caption{Qualitative comparison between ground truth, 3DGS, CompGS (16k), and \modelname\ (16k) (ours).}
\label{fig:qualitative_allscenes}
  \vspace*{-.1in}
\end{figure*} 
\paragraph{Qualitative results}
A qualitative comparison was conducted among ground truth, 3DGS, CompGS~(16k) and our \modelname~(16k), as illustrated in \cref{fig:qualitative_allscenes}. Despite utilizing only approximately $2.2\%$ of the memory, \modelname efficiently reconstructs scenes with high visual fidelity as 3DGS. Both 3DGS, CompGS~(16k), and ours~(16k) exibit limitations in capturing the fine details of ground in flower and treehill scenes.  
Another comparison, presented in \cref{fig:qualitative_0.5k}, underscores the advantages of \modelname over the best SP-based compression method, CompGS, at extremely low bitrates `0.5k', where the parameters are set to $K_s = K_r = 512, K_c= K_{sh} = 128$. The inability of CompGS to accurately reconstruct sharp details at such low bitrates is likely due to the STE solution for gradient collapse, which merely copies the gradients during training while disregarding quantization effects. These comparisons highlight the performance of \modelname in maintaining reconstruction quality, even under tight compression constraints. \looseness-1

\section{Conclusion and Discussion}
\label{sec:discussion_conclusion}
In this paper, we proposed \modelname, a novel VQ-based model for GS compression. The introduced NSVQ-based technique addresses the challenge of \textit{gradient collapse}, which arises from the inherent inconsistency between the discrete nature of quantization and gradient-descent optimization applied to Gaussian splat features. Our model achieves efficient compression of Gaussian splatting data while maintaining high reconstruction quality, as shown by both quantitative and qualitative evaluations. Furthermore, the streamlined storage structure enhances rendering speed, ensures compatibility with other compaction methods, and preserves the potential for broad industrial applications of 3DGS. It is worth stressing that while some ML-based methods achieve higher compression rates, the trained models lose appealing properties associated with 3DGS, \eg, real-time rendering and model editing capabilities which requires explicit modelling. Thus, we consider advancing SP-based GS compression methods to be an impactful direction for future research.
While \modelname\ demonstrates advancement, there remains potentials to further improve the compression ratio. One promising direction is to compress unquantized Gaussian features, \eg, quantizing spatial coordinates using space-filling curves. Additionally, the development of compaction models incorporating advanced heuristics could yield even greater compression efficiency.\looseness-1

A reference implementation of the methods is available at \url{https://github.com/AaltoML/NSVQGS}.

\begin{credits}
\subsubsection{\ackname}
This work was supported by the Research Council of Finland (362408, 339730) and the Finnish Center for Artificial Intelligence FCAI. We acknowledge the computational resources provided by the Aalto Science-IT project and CSC -- IT Center for Science, Finland.
\end{credits}

\clearpage

\bibliographystyle{splncs04}
\end{document}